\documentclass[lettersize,journal]{IEEEtran}
\usepackage{amsmath,amsfonts}
\usepackage{algorithmic}
\usepackage{algorithm}
\usepackage{array}
\usepackage[caption=false,font=normalsize,labelfont=sf,textfont=sf]{subfig}
\usepackage{textcomp}
\usepackage{stfloats}
\usepackage{url}
\usepackage{verbatim}
\usepackage{graphicx}
\usepackage{cite}
\usepackage{multirow}
\usepackage{orcidlink}
\usepackage{bbding}
\begin{document}
	

\title{Multimodal Feature Fusion Network with Text Difference Enhancement for Remote Sensing Change Detection}
\author{Yijun Zhou\textsuperscript{\orcidlink{0009−0005−4054−1594}}, Yikui Zhai\textsuperscript{\orcidlink{0000-0003-0154-9743}},~\IEEEmembership{Senior Member, IEEE}, Zilu Ying\textsuperscript{\orcidlink{0000−0002−3074−5586}}, Tingfeng Xian\textsuperscript{\orcidlink{0009-0002-8072-9725}},~\IEEEmembership{Student Member, IEEE}, Wenlve Zhou\textsuperscript{\orcidlink{0000-0002-7500-5581}}, Zhiheng Zhou\textsuperscript{\orcidlink{0000-0003-4040-0175}},~\IEEEmembership{Member, IEEE}, Xiaolin Tian, Xudong Jia\textsuperscript{\orcidlink{0000-0001-7911-8869}} , Hongsheng Zhang\textsuperscript{\orcidlink{0000-0002-6135-9442}},~\IEEEmembership{Senior Member, IEEE}, C. L. Philip Chen\textsuperscript{\orcidlink{0000-0001-5451-7230}},~\IEEEmembership{Life Fellow, IEEE}.
\thanks{This study was funded by Guangdong Basic and Applied Basic Research Foundation (No. 2021A1515011576), Guangdong Higher Education Innovation and Strengthening School Project (No. 2020ZDZX3031, No. 2022ZDZX1032, No. 2023ZDZX1029, 2024ZDZX1009), Wuyi University Hong Kong and Macao Joint Research and Development Fund (No. 2022WGALH19), Guangdong Jiangmen Science and Technology Research Project (No. 2220002000246, No. 2023760300070008390).(Yijun Zhou, Yikui Zhai, Zilu Ying, Tingfeng Xian, Wenlve Zhou contributed equally to this work.)(Corresponding author: Yikui Zhai and Hongsheng Zhang).}
\thanks{Yijun Zhou, Yikui Zhai, Zilu Ying and Tingfeng Xian are with the College of Electronics and Information Engineering, Wuyi University, Jiangmen, 529020, China(e-mail: 17346700814@163.com, yikuizhai@163.com, ziluy@163.com, tingfengxian21@163.com).}
\thanks{Wenlve Zhou, Zhiheng Zhou are with the School of Electronic and Information Engineering	and the Key Laboratory of Big Data and Intelligent Robot, Ministry of Education, South China University of Technology, Guangzhou, Guangdong 510641, China (e-mail: wenlvezhou@163.com; zhouzh@scut.edu.cn).}
\thanks{Xiaolin Tian are with the State Key Laboratory of Lunar and Planetary Sciences, Macau University of Science and Technology, Taipa, Macau (e-mail:xltian@must.edu.mo).}
\thanks{Xudong Jia is the College of Engineering and Computer Science, California State University, Northridge, 18111, America (e-mail: Xudong.Jia@csun.edu).}
\thanks{Hongsheng Zhang is with the Department of Geography, The University of Hong Kong, Hong Kong, China (e-mail: zhanghs@hku.hk).}
\thanks{C. L. Philip Chen is with the Faculty of Computer Science and Engineering, South China University of Technology, Guangzhou, Guangdong 510006, China (e-mail: philip.chen@ieee.org).}}

\markboth{Journal of \LaTeX\ Class Files,~Vol.~14, No.~8, August~2021}%
{Shell \MakeLowercase{\textit{et al.}}: A Sample Article Using IEEEtran.cls for IEEE Journals}


\maketitle

\begin{abstract}
Although deep learning has advanced remote sensing change detection (RSCD), most methods rely solely on image modality, limiting feature representation, change pattern modeling, and generalization—especially under illumination and noise disturbances. To address this, we propose MMChange, a multimodal RSCD method that combines image and text modalities to enhance accuracy and robustness. An Image Feature Refinement (IFR) module is introduced to highlight key regions and suppress environmental noise. To overcome the semantic limitations of image features, we employ a vision-language model (VLM) to generate semantic descriptions of bi-temporal images. A Textual Difference Enhancement (TDE) module then captures fine-grained semantic shifts, guiding the model toward meaningful changes. To bridge the heterogeneity between modalities, we design an Image-Text Feature Fusion (ITFF) module that enables deep cross-modal integration. Extensive experiments on LEVIR-CD, WHU-CD, and SYSU-CD demonstrate that MMChange consistently surpasses state-of-the-art methods across multiple metrics, validating its effectiveness for multimodal RSCD. Code is available at: https://github.com/yikuizhai/MMChange.
\end{abstract}

\begin{IEEEkeywords}
Remote sensing change detection (RSCD), Image Feature Refinement (IFR), Text Difference Enhancement (TDE), Image-Text Feature Fusion (ITFF).
\end{IEEEkeywords}

\section{Introduction}
\IEEEPARstart{R}{emote} sensing change detection (RSCD) involves utilizing remote sensing technology to analyze data gathered at various times in order to detect alterations in surface or environmental conditions. \cite{mandalEmpiricalReviewDeep2022a},\cite{ding2025survey}. RSCD has found broad applications in numerous fields, including land use change \cite{liCompetitionBiogeochemicalDrivers2023}, catastrophe evaluation \cite{zhengBuildingDamageAssessment2021}, urban development strategy \cite{yousifImprovingUrbanChange2013}, agricultural monitoring \cite{bullockTimelinessForestChange2022}, and ecological monitoring \cite{mucherLandCoverCharacterization2000}.

Traditional RSCD techniques can be classified into three primary types: algebraic approaches, transformation approaches, and classification approaches. Algebraic approaches rely on algorithmic operations, such as image differencing \cite{shiLandslideRecognitionDeep2021}, image ratio calculation \cite{zhangCoarsetoFineSemiSupervisedChange2018}, and image quantization \cite{daiQuantificationImpactMisregistration1997}, to generate difference features that creates change detection masks. However, image transformation techniques, including Principal Component Analysis (PCA) \cite{celikUnsupervisedChangeDetection2009c}, Change Vector Analysis (CVA) \cite{sahaUnsupervisedDeepChange2019b}, and Tasseled Cap Transformation (TCT) \cite{hanEfficientProtocolProcess2007}, reorganize spatial mappings to improve the visibility of change-related data. Finally, classification methods grounded in machine learning, such as Support Vector Machines (SVM) \cite{negriSpectralSpatialAware2021}, K-means clustering \cite{sunIterativeRobustGraph2021}, and Random Forest Regression \cite{seoFusionSARMultispectral2018}, utilize difference or spatiotemporal features to identify changes. Although these traditional methods can detect change regions to a certain extent, their susceptibility to noise, limited adaptability, and constrained classification accuracy impair their effectiveness in complex environments.

In the recent past, the accelerated growth of deep learning has emerged as a key driving force behind advancements in science and engineering, particularly revolutionizing the field of computer vision \cite{liu2021domain}, \cite{liu2023mscaf}, \cite{xie2024mifnet}, \cite{wang2023cross}, \cite{huang2021catfpn}. Deep learning-based RSCD methods, with their superior capabilities for feature learning and extraction, have outperformed traditional techniques, especially in managing complex scenes \cite{zhou2022spatial}. Convolutional Neural Networks (CNNs), known for their strong local feature extraction abilities, have gained remarkable results. Zhan et al. \cite{zhan2017change} put forward a deep Siamese convolutional neural network architecture for change detection using bitemporal remote sensing imagery. Daudt et al. \cite{cayedaudtFullyConvolutionalSiamese2018d} put forward three networks leveraging the UNet architecture \cite{ronnebergerUNetConvolutionalNetworks2015}. Wang et al. \cite{wang2023double} put forward a CD method integrating superpixels and CNNs for change detection in challenging scenes. However, CNNs have difficulty modeling long-range dependencies and understanding global context, which prevents them from fully leveraging spatial relationships in images. Consequently, some researchers have proposed Transformer models to address this deficiency. Since the advent of Transformers, their excellent performance has been widely recognized. The RSCD field has seen widespread use of transformers. Chen et al. \cite{chenRemoteSensingImage2022c} put forward a RSCD model based on Transformer architecture, which successfully modeled the spatiotemporal context between bitemporal images by combining Transformer architecture with semantic tokenization. Jiang et al. \cite{jiangVcTVisualChange2023b} propose VCT for CD, modeling structured info with GNNs (pixels as nodes), mining Top-K tokens via clustering, using self-cross-attention and APA to enhance features, and generating detailed change maps via prediction head. Wang et al. \cite{wangSummatorSubtractorNetwork2024d}put forward a novel Summator–Subtractor network for change detection, which effectively captures subtle differences in both the spatial and channel domains of bi-temporal images.

Existing methods primarily focus on unimodal change detection, neglecting the potential of multimodal approaches which could address issues like limited information representation and poor robustness \cite{saidi2024deep}, \cite{dongChangeCLIPRemoteSensing2024a}, \cite{liu2016deep}. Recently, multimodal learning has surfaced as a major topic of research. Dong et al. \cite{dongChangeCLIPRemoteSensing2024a} were the first to put forward a multimodal visual-text-based RSCD model, ChangeCLIP. This method inputs bitemporal remote sensing images into CLIP \cite{radfordLearningTransferableVisual2021}, converting images into text descriptions (category labels) and leveraging visual-language learning for change detection. While ChangeCLIP demonstrates the effectiveness of visual-language models in RSCD, several areas warrant further exploration and improvement. One limitation of ChangeCLIP is that its text prompts consist solely of high-level category labels, which offer too little detail to guide fine-grained change localization. In RSCD tasks, the key challenge is accurately locating the change areas in bitemporal images. Therefore, the text descriptions input into the model should be more precise, focusing on specific features of the change areas rather than merely pointing to the target category. This allows the model to more efficiently identify and grasp the details of the changes. A further drawback is that remote sensing images often suffer from noise, degradation, and varying environmental conditions, which hampers reliable change detection when prompted only by coarse labels. Thus, refining image features during the image processing phase aids the model in better extracting image features and provides higher-quality image feature for subsequent multimodal fusion, thereby improving detection accuracy and robustness. Finally, efficiently integrating multimodal information is crucial to leveraging the advantages of both visual and textual features. Since remote sensing images and text descriptions each present different forms of information and semantic levels, how to effectively combine the two to complement each other remains a significant challenge in multimodal learning. Particularly in RSCD tasks, images provide spatial details, while text can supplement semantic information. An effective fusion strategy can ensure spatial feature accuracy while enhancing the model's semantic understanding of change areas.

To address the aforementioned challenges, our research motivation is examined from three perspectives. First, most current remote sensing change detection (RSCD) methods rely on the image modality for change identification. However, image features are limited in expressing semantic hierarchies and handling complex scene variations, making it difficult to comprehensively capture the deep-level information of change regions. In addition, remote sensing images are prone to noise interference and illumination variations during the imaging process, which further undermines the robustness of these models and negatively impacts overall detection performance. Second, although some studies have introduced the text modality to improve change detection results, existing methods mostly rely on static or predefined text category labels, lacking detailed semantic expressions for specific change regions and failing to fully leverage the supplementary role of the text modality in the model’s semantic understanding. Third, current multimodal feature fusion strategies are not perfect, making it difficult to achieve deep collaboration and complementary utilization of image and text information, which limits the model’s ability to accurately model change regions. Therefore, we attempt to construct a more expressive multimodal fusion mechanism while improving the accuracy of text semantic guidance and the robustness of image features to enhance the model’s sensitivity to and understanding of changes in remote sensing images.

Based on the three research motivations outlined above, we propose a novel visual-language RSCD model named MMChange. Unlike existing unimodal RSCD approaches, MMChange integrates both image and text modalities to achieve more efficient change recognition. In terms of architecture, we introduce the Image Feature Refinement (IFR) module to enhance the representation capability of bitemporal images, thereby improving the model’s robustness to noise and illumination variations, and design the Text Difference Enhancement (TDE) module to enable precise semantic localization of change regions. Finally, the Image-Text Feature Fusion (ITFF) module is employed to effectively integrate multimodal information, thereby improving the accuracy and robustness of change detection. The main contributions are as follows:

\begin{enumerate}
	\item A multimodal RSCD model, MMChange, is proposed, which is based on visual-language model (VLM) and integrates both image and text modalities. Through this integration, the performance of MMChange has been significantly enhanced, surpassing the existing leading methods.
	\item A Text Difference Enhancement module (TDE) is introduced, which effectively enhances textual differences by processing textual descriptions. This leads to a better capture of textual features in areas with changes by the model. According to what we know, TDE is the first module to implement text difference enhancement to multimodal RSCD tasks.
	\item An Image Feature Refinement module (IFR) is designed, which strengthens the model's capability to gain image features, offering superior features for subsequent multimodal fusion, thereby boosting the accuracy of RSCD.
	\item An Image-Text Feature Fusion module (ITFF) is proposed, which integrates image and text features from IFR and TDE, exploiting semantic relationships and complementary information between image and text modalities, thereby enhancing accuracy and robustness of change detection.
\end{enumerate}
\section{Relate Works}
\subsection{Single-Modality Change Detection}
In contemporary mainstream change detection (CD) networks, most models continue to depend on single-modality data as input. The classification of these models consists of two main types: CNNs architectures and Transformer architectures. CNNs CD networks, known for their robust feature extraction capabilities, have shown exceptional performance in RSCD tasks. For instance, Zhang and Shi \cite{zhangFeatureDifferenceConvolutional2020} utilized VGG16 \cite{simonyanVeryDeepConvolutional2015} as a pretrained model to construct feature differences. By leveraging pretrained models, subtle changes in images are effectively captured, resulting in more accurate and reliable change features. Zhan et al. \cite{zhanChangeDetectionBased2017a} Put forward a incorporating a weighted contrastive loss to train the CD network and enable feature extraction directly from image pairs. Although CNNs-based CD methods have attained considerable success in certain application scenarios, their limitations in capturing contextual information still constrain their performance. CNNs primarily rely on local receptive fields and fixed convolutional kernels for feature extraction, this makes it hard to capture global context in intricate scenes. In recent years, Vision Transformers (ViT) \cite{hanSurveyVisionTransformer2023},\cite{leeMPViTMultiPathVision2022},\cite{zhuBiFormerVisionTransformer2023a} have been increasingly used for RSCD tasks by virtue of their powerful capability to model long-range global relationships within images. For instance, Zhang et al. \cite{zhangSwinSUNetPureTransformer2022c} utilized Swin Transformer blocks \cite{liuSwinTransformerHierarchical2021b} pure Transformer method with a U-shaped architecture was designed to construct an encoder-decoder structure. This design effectively leverages the Swin Transformer's robust feature extraction capabilities while facilitating efficient information transfer and fusion through the connected U-shape structure. Wu et al. \cite{wuCSTSUNetCrossSwin2023a} put forward a cross-Swin Transformer CD method that first employs ResNet \cite{heDeepResidualLearning2016b} to acquire features at different scales, followed by the computation of multi-scale difference features using a difference feature extraction module. By comparing and analyzing features at various scales, the network generates more precise and robust change features. Zhang et al. \cite{zhangMultiscaleCascadedCrossAttention2024a} put forward a multi-scale cascading cross-attention hierarchical network that combines small kernel convolutions stacked into convolutions with larger kernels, with efficient Transformers as the backbone. This structure enables the extraction and fusion of both local details and global contextual information, boosting the model's ability to capture extended dependencies within intricate scenes.
 
Although single-modal change detection networks excel in feature learning, their performance in complex scenarios is limited by the inability to leverage multi-source information, resulting in insufficient robustness. For example, CNNs struggle with long-range dependencies and spectral confusion, while Transformers lack semantic support for scenario understanding (e.g., "farmland-to-factory" changes). Additionally, single-modal models are vulnerable to noise (e.g., clouds, sensor errors) without complementary information. Multimodal methods address these gaps by fusing visual and textual features, enhancing spatial-semantic complementarity and detection reliability in complex contexts.
\subsection{Multimodal Representation Learning}
As large models in deep learning rapidly progress and artificial intelligence, the limitations of single-modality feature representations in addressing complex downstream tasks have become increasingly evident. Although traditional single-modality learning methods have achieved notable success in specific domains, they often depend on data from a single modality for task reasoning and decision-making \cite{yingDGMA2NetDifferenceGuidedMultiscale2024},\cite{yingDSHyFANetDeeplySupervised2024a},\cite{zhaiCASNetComparisonBasedAttention2024d},\cite{zhaiEfficientAdjacentFeature2024}. However, in many real-world applications, a single data source is inadequate to capture the rich information necessary for tasks, particularly when confronted with diverse input data and intricate scenarios. In such instances, model performance is frequently severely constrained. For example, in fields like autonomous driving, remote sensing image analysis, and sentiment analysis, information typically comes from different modalities such as vision, speech, text, and sensor data. A single modality alone cannot effectively integrate these diverse sources of information, leading to difficulties in making accurate judgments and decisions. To overcome this issue, an increasing number of researchers have started to propose and explore multimodal representation learning. By integrating data from different modalities, multimodal learning can simultaneously leverage the complementarity and interrelationships between multiple information sources, consequently enhancing the model's capacity to reason through and manage complex tasks. For instance, Li et al. \cite{li2025urbansam} proposed UrbanSAM to address the heterogeneity of multimodal remote sensing data. The method designs adapters based on multiresolution analysis and fuses features through a cross-attention mechanism, improving urban land cover segmentation performance on global multimodal datasets without requiring manual prompts and effectively adapting to multiscale variations. Through joint learning of multimodal data, models not only improve performance on specific tasks but also enhance their adaptability to unknown and complex situations. Thus, multimodal representation learning has become a crucial research direction in artificial intelligence, driving breakthroughs in model capabilities and offering new technical pathways for the fusion of diverse data sources and intelligent decision-making in practical applications.

However, research on multimodal-based RSCD remains relatively limited. The primary reason for this is that existing RSCD algorithms typically focus on processing data from a single modality, emphasizing image modality features while neglecting text modality features, and they lack an effective multimodal fusion method. As a result, these approaches struggle to fully leverage the potential of multimodal data. Although Dong et al. \cite{dongChangeCLIPRemoteSensing2024a} proposed the first multimodal RSCD model, ChangeCLIP, which integrates multimodal information by using target categories as textual prompts, this approach relies on simplistic text prompts and may not effectively guide the model in seizing the complex associations between images and text. In order to tackle the shortcomings of current methods, we put forward an innovative multimodal RSCD model based on VLM, called MMChange. This model combines multimodal information to fully exploit the complementarity between remote sensing images and text, enhancing the accuracy and robustness of RSCD. MMChange consists of three core modules: First, the IFR refines the features of remote sensing images. Second, the TDE Module strengthens key information related to changes in text descriptions, helping the model more accurately understand features of changes. Finally, the ITFF effectively fusion image and text information, enabling the model to fully utilize the correlations between the two, consequently boosting the precision of change detection.
\section{Methods}
In this section, we outline the overall structure of MMChange in Subsection A. To enhance text descriptions and emphasize key features of change areas, we present the TDE module in Subsection B. To refine image features, we propose the IFR module in Subsection C. Finally, to improve the fusion of multimodal features, we introduce the Image-Text Feature Fusion (ITFF) module in Subsection D.
\subsection{Overall Structure of MMChange}
\begin{figure*}
	\centering
	\includegraphics[width=1\textwidth]{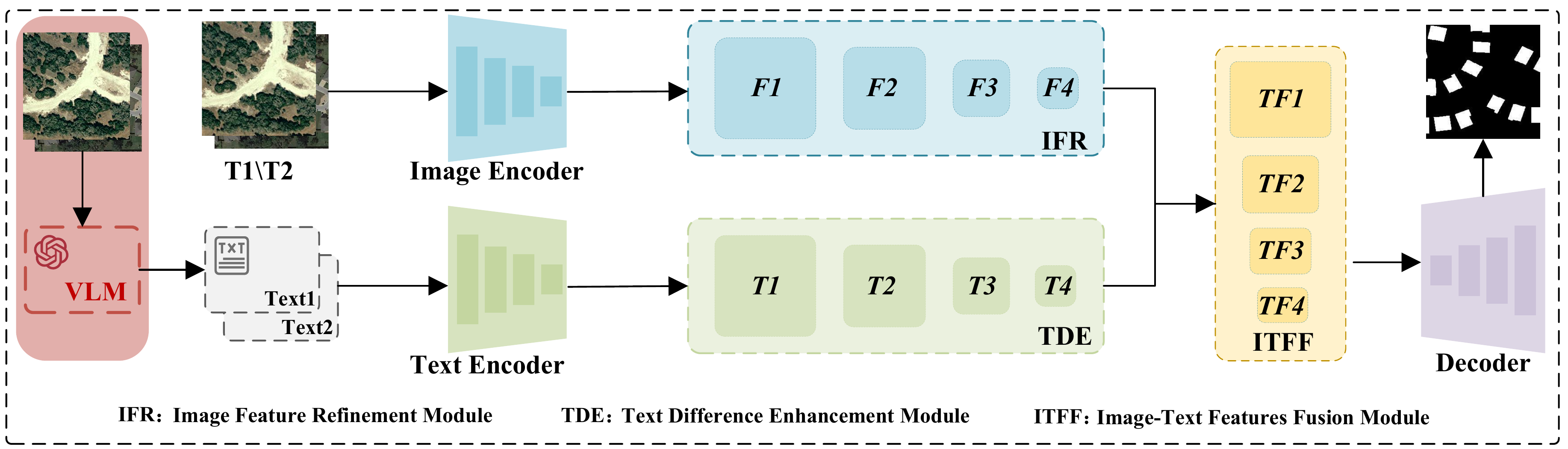}
	\caption{illustrates the architecture of the MMChange model, which is divided into two image and text stages. In the image stage, the image encoder generates four-scale features, and the IFR module is utilized to enhance the bitemporal remote sensing image features at four scales (F1, F2, F3, F4). In the text stage, the TDE module amplifies the differences between the text descriptions of bitemporal images at four scales (T1, T2, T3, T4). Finally, the ITFF module fuses image and text features (e.g., image feature F1 and text feature T1 are input into TF1, and so on for other scales), and the fused four-scale features are fed into the decoder to generate the final change mask.}
	\label{Fig 1: MMChange_framework}
\end{figure*}

\begin{figure}
	\centering
	\includegraphics[width=3.5in]{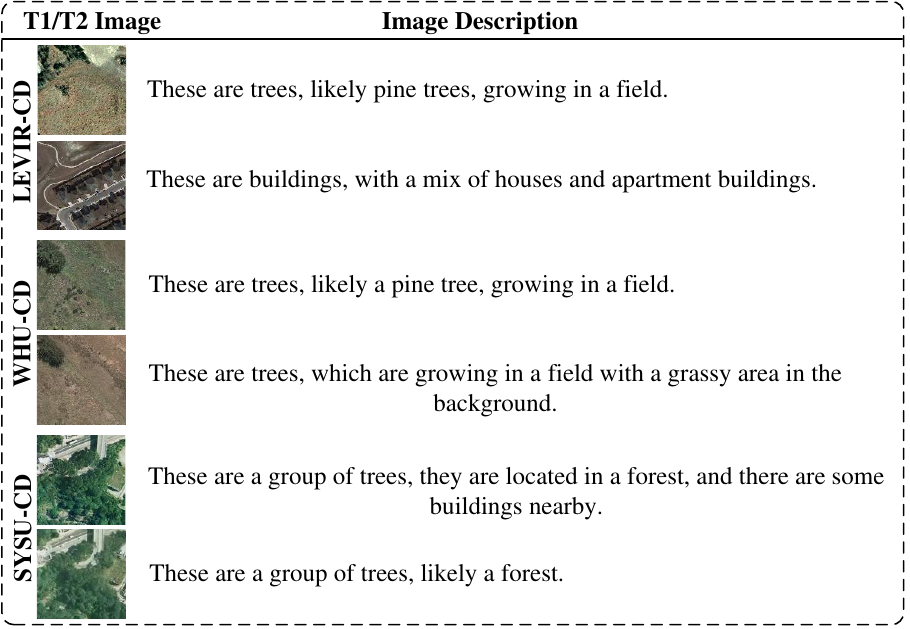}
	\caption{illustrates the textual descriptions of the three datasets (LEVIR-CD, WHU-CD, and SYSU-CD) generated by the VLM (TinyLLaVA).}
	\label{Fig +: Text}
\end{figure}

In this paper, we put forward an RSCD method leveraging VLM, named MMChange, as illustrated in Fig.~\ref{Fig 1: MMChange_framework}. Specifically, the bitemporal RS images are initially processed through an image encoder based on the ResNet50 architecture \cite{heDeepResidualLearning2016b} for preliminary feature extraction. These features are subsequently refined using the IFR module, which enhances attributes such as shape, contour, and texture, resulting in more comprehensive image features that facilitate accurate change detection. Following this, the VLM (TinyLLaVA) \cite{zhou2024tinyllava} generate textual descriptions of bitemporal images. Specifically, the prompt fed into TinyLLaVA is "What are the components in this picture?". When generating text with TinyLLaVA, we used its default parameter configuration (e.g., temperature = 0.2, top\_p = 0.9, etc.), as illustrated in Fig.~\ref{Fig +: Text}, TinyLLaVA is a VLM specifically designed for small-scale multimodal tasks, featuring only a few hundred million parameters and low computational overhead. It can efficiently generate accurate text descriptions for bi-temporal remote sensing images and excels in cross-modal alignment, capable of mapping visual features to semantic differences. Its small parameter count and low computational cost provide reliable support for subsequent text difference enhancement, which are then input into the CLIP text encoder \cite{radfordLearningTransferableVisual2021} to obtain textual features. The TDE module is utilized to enhance the discrepancies between two text representations, effectively capturing semantic variations in bitemporal image features. This text-difference-based feature enhancement allows the model to more precisely localize and describe changed areas, thereby improving detection accuracy. Moreover, by highlighting semantic variations between bitemporal descriptions, the TDE module enhances the discriminative power of the textual features, enabling the subsequent fusion stage (ITFF) to better align semantic cues with visual differences. Finally, the multimodal features are integrated and sent into the decoder to create a change mask.
\subsection{Text Difference Enhancement Module (TDE)}
In multimodal RSCD, the TDE module is designed to emphasize variations in the textual modality, consequently increasing the precision and robustness of CD. Textual information in RS images, such as land use, roads, and buildings, often contains rich semantic features that effectively indicate areas of change. Fig.~\ref{Fig +: Text} illustrates several textual descriptions generated by the vision-language model (TinyLLaVA) on three remote sensing change detection datasets. For example, the model generates similar descriptions for multiple bi-temporal image pairs, such as “These are trees, likely a pine tree, growing in a field” and “These are a group of trees, likely a forest,” indicating semantic redundancy and a lack of fine-grained details when distinguishing between different scenes and temporal changes. Additionally, for regions containing prominent building structures, the generated description—“These are buildings, with a mix of houses and apartment buildings”—captures only the basic scene semantics while failing to precisely convey the change areas or types of changes. These examples reveal the inherent limitations of the raw textual modality, including ambiguity, semantic inconsistency, and insufficient detail, thereby highlighting the necessity of our proposed TDE module in improving semantic expressiveness and enhancing the accuracy of change region discrimination. The TDE module employs scaled dot-product attention \cite{vaswani2017attention} by introducing multi-scale information aggregation and salient region guidance mechanisms, the model can adaptively capture key regions with significant differences between bi-temporal remote sensing images, highlighting the semantic features of change areas during text generation. Compared with traditional attention mechanisms, this approach enhances spatial perception of change regions, thereby improving the sensitivity and discriminative power of the generated text to change patterns, and achieving more effective textual difference enhancement. this not only boosts the discriminability and robustness of textual features but also generates rich semantic cues that, when passed into the fusion block alongside fine‐grained spatial features from the visual branch, guide cross‐modal attention to semantically meaningful regions, ensuring that visual and textual representations mutually inform and amplify each other’s strengths to achieve higher overall precision in change detection.

\begin{figure}
	\centering
	\includegraphics[width=3.5in]{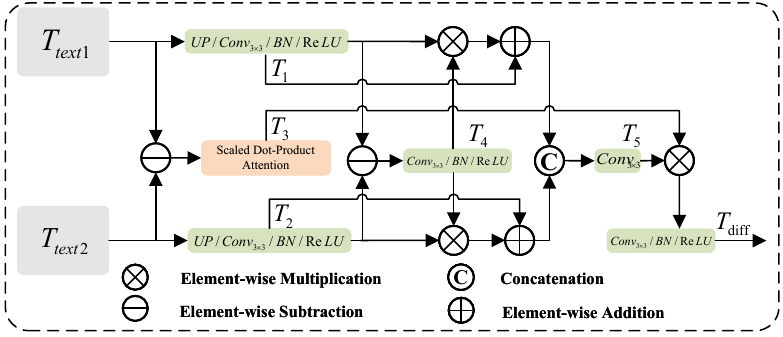}
	\caption{TDE Structure.}
	\label{Fig 3: TDE module}
\end{figure}

In the text stage of the MMChange model, we utilize the TDE, as illustrated in Fig.~\ref{Fig 3: TDE module}. The primary aim of the TDE is to enhance the differential features between the textual descriptions extracted from bitemporal remote sensing images, thereby assisting the model in recognizing and analyzing change regions. Specifically, take the first scale as an example, the textual features, $T_{text1}$ and $T_{text2}$, extracted by the text encoder from the bitemporal RS images undergo a series of operations, including upsampling, convolution, normalization, and activation, to derive features $T_{1}$ and $T_{2}$. To further explore the textual difference features between the bitemporal images, we subtract $T_{text1}$ and $T_{text2}$ and apply scaled dot-product attention to generate feature $T_{3}$. The scaled dot-product attention mechanism demonstrates significant advantages in enhancing textual feature differences by improving attention focus, enhancing feature representation, boosting the model’s generalization ability, optimizing computational efficiency, and facilitating cross-modal information fusion. This allows it to more precisely capture and amplify subtle differences between textual descriptions of bitemporal remote sensing images. Subsequently, a series of convolution, normalization, activation, and other operations are applied to generate features $T_{4}$ and $T_{5}$. Next, we perform element-wise multiplication between features $T_{3}$ and $T_{5}$. This operation not only integrates the advantages of both features but also preserves their unique information. To boost the ability of the fused features to represent information, we apply another convolution operation. The convolution operation captures the local correlations between features, resulting in a more enriched and discriminative final feature representation, $T_{diff}$. The TDE process can be described as follows:
\begin{equation}T_1=\operatorname{Re}LU(BN(Con\nu_{3\times3}(UP(T_{text1}-T_{text2}))))\end{equation}
\begin{equation}T_2=\operatorname{Re}LU(BN(Con\nu_{3\times3}(UP(T_{text1}-T_{text2}))))\end{equation}
\begin{equation}T_3=SDPA(T_{text1}-T_{text2})\end{equation}
\begin{equation}T_4=\operatorname{Re}LU(BN(Con\nu_{3\times3}(T_1-T_2)))\end{equation}
\begin{equation}T_5=Conv_{3\times3}(Cat(T_1+(T_1\cdot T_4),T_2+(T_2\cdot T_4)))\end{equation}
\begin{equation}T_{diff}=\operatorname{Re}LU(BN(Conv_{3\times3}(T_5\cdot T_3)))\end{equation}
where SDPA($\cdot$) refers to the scaled dot-product attention operation.

\subsection{Image Feature Refinement Module (IFR)}
In multimodal RSCD, image features are refined to enhance their clarity and prominence, providing high-quality features for subsequent multimodal fusion. The IFR model first integrates coordinate and channel information, which improves the model’s ability to obtain image features and enhances the recognition of object location, shape, and semantic information. The refined features not only suppress noise interference but also enhance low-level spatial cues (e.g., edges and textures), thereby strengthening their complementarity with textual features and providing reliable anchors for precise image–text alignment and fusion, which in turn improves detection robustness and consistency in complex scenarios. Furthermore, the feature map is divided into distinct groups, with each group independently processing different parts or frequencies of the image. Within each group, features are further refined using residual learning and convolution operations. This procedure boosts the precision and robustness of change detection. Therefore, feature refinement establishes a solid foundation for multimodal fusion and markedly improves the performance of CD.
\begin{figure}
	\centering
	\includegraphics[width=3.5in]{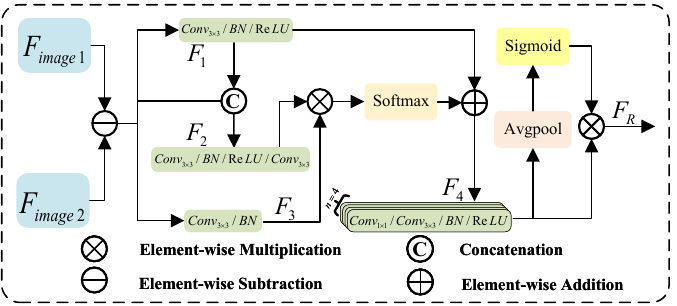}
	\caption{IFR Structure.}
	\label{Fig 2: IFR module}
\end{figure}

The IFR module is depicted in Fig.~\ref{Fig 2: IFR module}. Specifically, take the first scale as an example, the bitemporal image features, $F_{image1}$ and $F_{image2}$, extracted by the image encoder are subjected to subtraction, and the resulting difference is processed through convolutional layers, normalization layers, and activation functions to generate a new feature, $F_{1}$. Additionally, the difference between the two features is concatenated with $F_{1}$, and after undergoing similar processing, another new feature, $F_{2}$, is produced through convolutional layers. Furthermore, the difference undergoes convolutional processing. and normalization layers to create a feature, $F_{3}$. The convolutional layers reduce the number of training parameters by utilizing shared parameters, consequently boosting the model’s capacity to understand translational invariance. The normalization layers standardize the features, constraining the parameter distribution across layers and reducing model complexity, which in turn improves the model's universality ability. By incorporating activation layers, non-linearity is introduced, allowing the model to learn and approximate intricate non-linear functions. Subsequently, $F_{2}$ and $F_{3}$ are multiplied, followed by a softmax operation, and then added to $F_{1}$. The feature map is divided into distinct groups (we are divided into four groups, i.e., n=4), with each group capable of independently processing various image sections or frequency information. Within each group, residual learning and convolution operations are employed to refine the features and generate $F_{4}$. Finally, $F_{4}$ undergoes average pooling and a Sigmoid operation, which is then multiplied by $F_{4}$ to produce the refined feature, $F_{R}$. The IFR process can be summarized as follows:
\begin{equation}F_1=ReLU(BN(Con\nu_{3\times3}(F_{image1}-F_{image2})))\end{equation}
\begin{equation}F_2=Cat(F_1,(F_{image1}-F_{image2}))\end{equation}
\begin{equation}F_3=BN(Conv_{3\times3}(F_{image1}-F_{image2}))\end{equation}
\begin{equation}F_4=ReLU\!(BN\!(Con\nu_{3\times3}\!(Con\nu_{1\times1}\!(Softmax\!(F_2 \cdot F_3\!)))))_{n=4}\!\end{equation}
\begin{equation}F_R=Sigmoid(A\nu gpool(F_4))\cdot F_4\end{equation}

\subsection{Image-Text Feature Fusion Module (ITFF)}
In multimodal change detection, combining image and text features is key, as it effectively unites visual and textual information to enhance change perception. The image modality primarily provides spatial structure and visual features, while the text modality offers rich semantic information, including details about buildings, roads, and other critical elements necessary for detection. By merging these two modalities, the limitations of each individual modality can be effectively addressed, minimizing modality discrepancies and boosting the model’s capability to recognize changes, thereby enhancing both the precision and robustness of CD. To capture the features of both image and text, the ITFF module effectively integrates features from both modalities through various attention mechanisms. To begin with, channel attention concentrates on the feature map’s channel dimension, automatically strengthening channels critical for change detection and suppressing irrelevant or noisy ones to enhance the representation of key information. Then, spatial attention focuses on image regions, guiding the model to identify and prioritize areas with substantial changes while disregarding regions with minimal or no change. Finally, pixel attention refines this process at the finest granularity, precisely capturing subtle yet significant alterations at the individual pixel level. Through this multi-level feature extraction, the ITFF module not only boosts the fusion of image and text features but also effectively mitigates the limitations of each modality, leading to more accurate and comprehensive CD and boosting the model's ability to perceive complex changes.

We input the refined image features and the enhanced differential textual features into the ITFF module for processing, as illustrated in Fig.~\ref{Fig 4: ITF module}. The ITFF module adeptly integrates various attention mechanisms to analyze and capture from both image and text features. By amalgamating different types of attention mechanisms, the module can more effectively extract feature from both image and text. By focusing on the channel dimension of feature maps, channel attention enhances critical channels and suppresses irrelevant ones. Spatial attention focuses on the spatial positions within the feature map, identifying key regions. Pixel attention, on the other hand, focuses on individual pixel-level features, allowing for precise capture of detailed information. This multi-level feature extraction makes the fused feature representation richer and more accurate. Through this series of multi-level and multi-dimensional feature extraction and fusion, the feature representation generated by the ITFF module becomes more comprehensive and accurate, providing strong support for subsequent change detection tasks. 

In practical implementation, take the first scale as an example, we first perform an element-wise addition of the image features F and text features T to obtain the initial fused representation TF. To enhance the modeling capacity for key semantics and salient regions, we incorporate both channel attention and spatial attention mechanisms into the fused multimodal feature TF. In the spatial attention module, we compute the average and maximum values along the channel dimension of the TF feature map, resulting in two spatial descriptors $S_{1}$ and $S_{2}$, which capture different types of activation responses. These are concatenated and passed through a convolutional layer followed by a sigmoid activation to generate the spatial attention map, thereby enhancing the model's focus on critical spatial regions.For the channel attention module, we perform global average pooling along the spatial dimensions to produce a global channel descriptor $C_{1}$. This descriptor is processed by a convolutional layer and activation function to generate the channel attention weights, which emphasize more semantically informative channels. Next, in the pixel-level attention modeling stage, we add the outputs of the spatial and channel attention modules to integrate saliency guidance from different dimensions, achieving a more comprehensive feature enhancement. The result is reshaped to obtain a refined feature representation $P_{1}$, which is adapted for subsequent fusion operations. Simultaneously, the initial fused feature TF undergoes the same reshaping to produce $P_{2}$, ensuring dimensional consistency. We then concatenate $P_{1}$ and $P_{2}$ along the channel dimension to obtain the combined feature $P_{3}$. A convolutional layer followed by a sigmoid activation is applied to $P_{3}$, generating a pixel-level fusion attention map, which allows fine-grained response modulation at each spatial location. Compared with approaches relying solely on channel or spatial attention, the incorporation of pixel-level modeling further improves the model’s response accuracy and local discriminative capability during multimodal fusion. This mechanism not only enhances the granularity of feature representation but also strengthens the model’s ability to perceive object boundaries, fine structures, and local semantic differences, making it especially suitable for precise feature localization in complex scenes. Finally, the pixel-level attention map is multiplied with the original fused feature TF to obtain the attention-enhanced multimodal feature $P_{4}$. A final convolutional layer is then applied to integrate the multi-level attention information, producing the output feature $F_{out}$, which serves as the result of the ITFF module. This approach ensures that the ITFF module fully exploits and utilizes the latent information in both image and text, providing high-quality feature representations for RSCD tasks. The fused feature $F_{out}$ from the ITFF module is then input into the decoder \cite{liLightweightRemoteSensing2023} to generate the final change mask. The process of ITFF can be summarized as follows:
\begin{figure}[!t]
	\centering
	\includegraphics[width=3.5in]{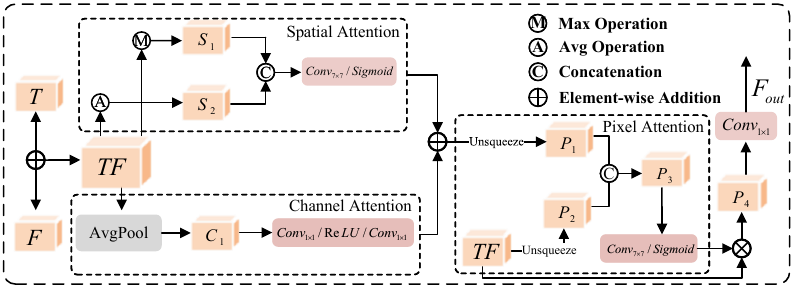}
	\caption{ITFF Structure.}
	\label{Fig 4: ITF module}
\end{figure}
\begin{equation}TF=T+F\end{equation}
\begin{equation}F_{out}=TF\cdot Sigmoid(PA(CA(TF)+SA(TF)))\end{equation}
where PA() indicates the pixel attention operation, CA() indicates the channel attention operation, and SA() indicates the spatial attention operation.
\section{Experiments}
To assess the effectiveness of the MMChange model, we carried out an evaluation using three datasets. We performed a comparison of our model with the latest state-of-the-art (SOTA) methods on these datasets. The experiments' findings clearly confirm the effectiveness of MMChange. We offer a detailed explanation of the experimental setup and evaluation metrics, and we utilize result visualization to enhance the intuitiveness of the outcomes, thereby effectively showcasing the effectiveness of our model across different datasets and comparing it with other methods. Finally, ablation experiments were performed to provide additional validation for the effectiveness of MMChange.
\subsection{Data Description}
We introduce three well-known datasets, LEVIR-CD, WHU-CD, and SYSU-CD, used to assess the performance of MMChange.
\begin{enumerate}
	\item LEVIR-CD \cite{chenSpatialTemporalAttentionBasedMethod2020b}: It comprises 637 pairs of bitemporal RS images, each sourced from Google, with a resolution of 1024 $\times$ 1024 pixels and a spatial resolution of 0.5 meters per pixel.
	\item WHU-CD \cite{jiFullyConvolutionalNetworks2019a}: It consists of two aerial images, each with a resolution of 32,507 $\times$ 15,354 pixels and a spatial resolution of 0.3 meters per pixel. We divided the aerial images into 7,620 image patches, each measuring 256 $\times$ 256 pixels.
	\item SYSU-CD \cite{shiDeeplySupervisedAttention2022b}: It contains 20,000 pairs of aerial images, taken over Hong Kong between 2007 and 2014, with each image having a resolution of 256 $\times$ 256 pixels and a spatial resolution of 0.5 meters per pixel.
\end{enumerate}
\subsection{Evaluation Metrics}
To thoroughly assess the effectiveness of MMChange in the CD task, we employed four widely recognized metrics: Precision (P), Recall (R), Intersection over Union (IOU), and F1 score (F1). In RSCD tasks, IOU measures the overlap between predicted and actual change regions, while F1 combines precision and recall, effectively addressing class imbalance. Together, they provide a comprehensive evaluation of the model's localization accuracy and detection capability \cite{wangSummatorSubtractorNetwork2024d}, \cite{yingDGMA2NetDifferenceGuidedMultiscale2024}. Consequently, we utilized these two metrics as the primary criteria for assessing the model's overall performance. Importantly, when comparing our model with others, we ensured that all training parameters were set to their default values to maintain fairness and comparability in the experiments. The calculations for each metric are as follows:
\begin{equation}
	\label{eq25}
	Precision=\frac{T_{P}}{T_{P}+F_{P}}
\end{equation}
\begin{equation}
	\label{eq26}
	Recall=\frac{T_{P}}{T_{P}+F_{N}}
\end{equation}
\begin{equation}
	\label{eq27}
	IOU=\frac{T_{P}}{T_{P}+F_{N}+F_{P}}
\end{equation}
\begin{equation}
	\label{eq28}
	F1=2\times\frac{Precision \times Recall}{Precision+Recall}
\end{equation}
where $T_P$ indicates the total number of pixels correctly identified as changed, $F_P$ indicates the total number of pixels that are actually unchanged but incorrectly labeled as changed, and $F_N$ refers to the total number of pixels that actually have changed but are incorrectly labeled as unchanged.
\subsection{Implementation Details}
MMChange, constructed with PyTorch, is trained on an NVIDIA Titan RTX (24GB) GPU, leveraging the Adam optimizer with a momentum of 0.9 and a weight decay factor of 0.0001. Additionally, the values for $\beta_{1}$, $\beta_{2}$ are set to 0.9 and 0.99, respectively. To effectively adjust the learning rate, we employ a polynomial decay scheme, where the learning rate is modified according to the formula $(1-(cur\_iteration/\max\_iteration))^{power}\times lr$, with the parameters power and $max\_iteration$ set to 0.9 and 40,000. The initial learning rate is set to 0.0005, with a batch size of 32. We use random flipping, cropping, and temporal random swapping as data augmentation techniques for the input images.
\subsection{Comparison with State-of-the-Art Models}
In this study, MMChange is compared with twelve SOTA RSCD models, specifically including FC-Conc \cite{cayedaudtFullyConvolutionalSiamese2018d}, FC-Diff \cite{cayedaudtFullyConvolutionalSiamese2018d}, BIT \cite{chenRemoteSensingImage2022c}, ICIF-Net \cite{fengICIFNetIntraScaleCrossInteraction2022c}, ChangeFormer \cite{bandaraTransformerBasedSiameseNetwork2022a}, VcT \cite{jiangVcTVisualChange2023b}, DMINet \cite{fengChangeDetectionRemote2023c}, SEIFNet \cite{huangSpatiotemporalEnhancementInterlevel2024c}, BiFA \cite{zhangBiFARemoteSensing2024a}, SSCD-CD \cite{wangSummatorSubtractorNetwork2024d}, DGMA2-Net \cite{yingDGMA2NetDifferenceGuidedMultiscale2024}, and ChangeCLIP \cite{dongChangeCLIPRemoteSensing2024a}. Notably, the IOU and F1 are key metrics that most effectively reflect model performance in RSCD tasks. Therefore, we assess the performance of each model using F1 and IOU metrics. For a fair comparison, we used the open-source code to re-implement these methods. \textcolor{red}{Furthermore, it should be noted that ChangeCLIP is a representative image-text-based model for RSCD. Therefore, comparing MMchange with ChangeCLIP ensures the fair evaluation of multimodal methods.}
\begin{table*}
	\caption{Quantitative Evaluation Results Of The LEVIR-CD, WHU-CD And SYSU-CD, Dataset Are Presented. The Best Values Are Highlighted In Bold, And All Values Are Expressed As Percentages.
		\label{tab1:ovaer_result}}
	\centering

	\renewcommand{\arraystretch}{1.2}	\resizebox{\textwidth}{!}{
		\begin{tabular}{cccc|cccc|cccc|cccc}
			\hline
			\multirow{2}{*}{Method} &\multirow{2}{*}{Params.(M)}&\multirow{2}{*}{FLOPs(G)}&\multirow{2}{*}{FPS}& \multicolumn{4}{c|}{\textbf{LEVIR-CD}}                                                           & \multicolumn{4}{c|}{\textbf{WHU-CD}}                                                     
			& \multicolumn{4}{c}{\textbf{SYSU-CD}} \\
			\cline{5-16}
			&&&& \multicolumn{1}{c}{IOU} & \multicolumn{1}{c}{F1} & \multicolumn{1}{c}{R} & \multicolumn{1}{c|}{P} & \multicolumn{1}{c}{IOU} & \multicolumn{1}{c}{F1} & \multicolumn{1}{c}{R} & \multicolumn{1}{c|}{P} & \multicolumn{1}{c}{IOU} & \multicolumn{1}{c}{F1} & \multicolumn{1}{c}{R} & \multicolumn{1}{c}{P}\\
			\hline
			FC-Conc \cite{cayedaudtFullyConvolutionalSiamese2018d}                 &1.54&5.33&460.22 & 81.38          & 89.74          & 87.42          & 92.18          & 58.45          & 73.77          & 75.95          & 71.72          & 60.05          & 75.04          & 78.38          & 71.98          \\
			FC-Diff \cite{cayedaudtFullyConvolutionalSiamese2018d}                 &1.34&4.72&470.80 & 82.21          & 90.23          & 88.14          & 92.43          & 59.59          & 74.68          & 68.60          & 81.95          & 51.65          & 68.12          & 59.98          & 78.81          \\
			BIT \cite{chenRemoteSensingImage2022c}                    &3.04&8.75&83.76 & 82.08          & 90.16          & 88.18          & 92.23          & 60.14          & 75.11          & 70.40          & 80.50          & 63.31          & 77.53          & 75.03          & 80.21          \\
			ICIF-Net \cite{fengICIFNetIntraScaleCrossInteraction2022c}                &25.30&23.84& 80.75& 81.33          & 89.70          & 88.72          & 90.71          & 75.05          & 85.75          & 79.25          & 93.40          & 62.44          & 76.88          & 72.95          & 81.26          \\
			ChangeFormer \cite{bandaraTransformerBasedSiameseNetwork2022a}            &41.02&202.78& 10.09& 81.98          & 90.10          & 88.92          & 91.31          & 84.74          & 91.74          & 88.36          & 95.39          & 60.91          & 75.7           & 71.15          & 80.88          \\
			VcT \cite{jiangVcTVisualChange2023b}                     &3.57&10.64&192.48 & 79.57          & 88.63          & 84.14          & \textbf{93.62} & 60.60          & 75.47          & 72.13          & 79.13          & 65.50           & 79.15          & 77.29          & 81.11          \\
			DMINet \cite{fengChangeDetectionRemote2023c}                  &6.24&14.55& 142.02 & 83.44          & 90.97          & 89.82          & 92.16          & 61.91          & 76.47          & 75.19          & 77.81          & 67.58          & 80.65          & 76.16          & 85.71          \\
			SEIFNet \cite{huangSpatiotemporalEnhancementInterlevel2024c}                 &27.91&8.37& 86.87& 81.78          & 89.98          & 88.28          & 91.74          & 65.87          & 79.42          & 72.55          & 87.73          & 67.90           & 80.88          & \textbf{83.59} & 78.35          \\
			BiFA \cite{zhangBiFARemoteSensing2024a}                    &5.58&53.01&23.74 & 82.39          & 90.35          & 89.19          & 91.54          & 88.94          & 94.15          & 93.12          & 95.20          & 71.39          & 83.31          & 80.83          & 85.94          \\
			SSCD-CD \cite{wangSummatorSubtractorNetwork2024d}                 &3.22&9.26& 221.16 & 82.53          & 90.43          & 89.68          & 91.19          & 65.74          & 79.33          & 73.00          & 86.86          & 64.25          & 78.23          & 76.11          & 80.47          \\
			DGMA2Net \cite{yingDGMA2NetDifferenceGuidedMultiscale2024}               &37.10&18.10&67.01 & 83.48          & 91.00          & 89.72          & 92.31          & 87.93          & 93.58          & 91.82          & 95.40          & 70.91          & 82.98          & 81.35          & 84.68          \\
			ChangeCLIP \cite{dongChangeCLIPRemoteSensing2024a}              &117.54&41.89&32.03 & 84.47          & 91.58          & 90.62          & 92.57          & 87.56          & 93.36          & 92.35          & 94.40          & 71.27          & 83.22          & 81.48          & 85.04          \\
			\hline
			Ours                    &42.90&113.52& 29.15 & \textbf{85.06} & \textbf{91.93} & \textbf{91.06} & 92.78          & \textbf{90.90} & \textbf{95.23} & \textbf{93.68} & \textbf{96.83} & \textbf{72.05} & \textbf{83.76} & 81.04          & \textbf{86.66} \\
			\hline    
	\end{tabular}}
\end{table*}

\begin{figure*}
	\centering
	\includegraphics[width=1\textwidth]{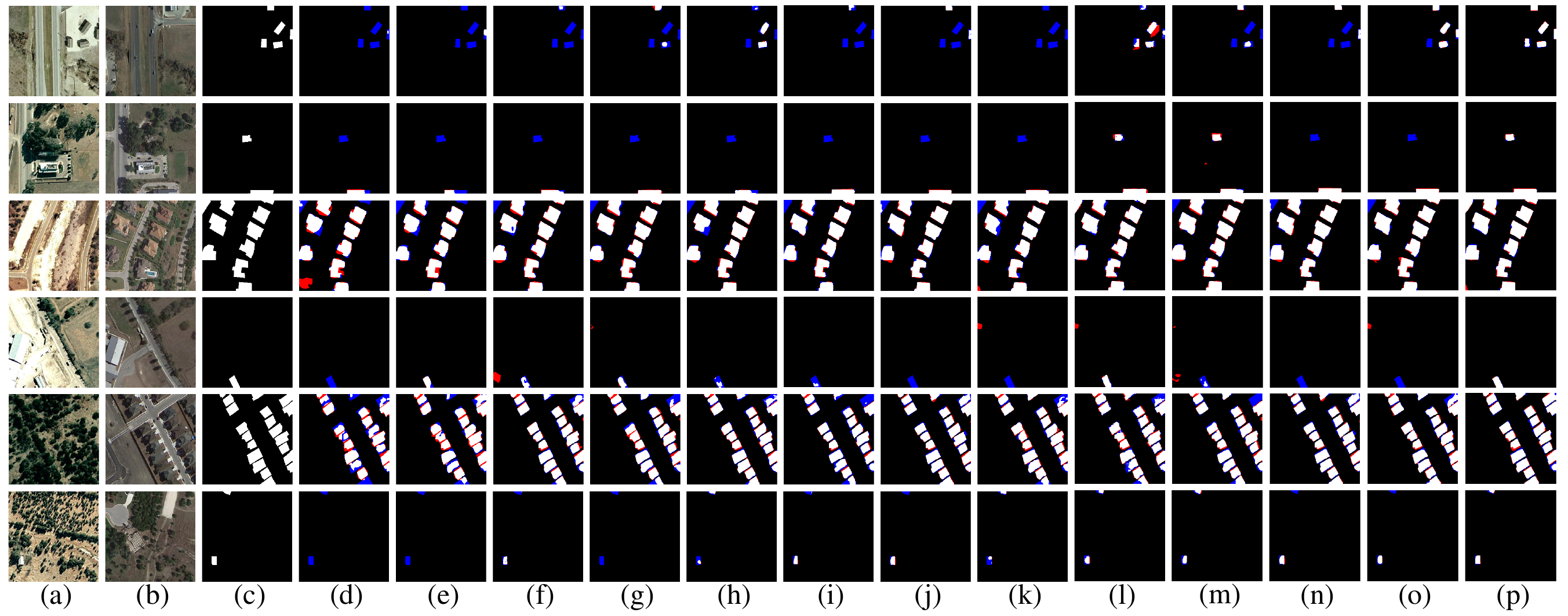}
	\caption{Visual comparative results of MMChange and other state-of-the-art methods on LEVIR-CD, where white denotes true positives, black denotes true negatives, blue denotes false negatives, and red denotes false positives. (a)T1 images, (b)T2 images, (c)Ground-truth, (d)FC-Conc, (e)FC-Diff, (f)BIT, (g)ICIF-Net, (h)ChangeFormer, (i)VcT, (j)DMINet, (k)SEIFNet, (l)BiFA, (m)SSCD-CD, (n)DGMA2Net, (o)ChangeCLIP, (p)MMChange.}
	\label{Fig 5: vis_levir}
\end{figure*}

Table ~\ref{tab1:ovaer_result} displays the experimental results for the LEVIR-CD. The ChangeCLIP model outperformed the other comparison models., with an IOU of 84.47\% and an F1 of 91.58\%. In comparison, the proposed MMChange model attained an IOU of 85.06\% and an F1 of 91.93\%, exceeding the ChangeCLIP model in both key metrics. Specifically, MMChange improved the IOU and F1 by 0.59\% and 0.35\%, respectively. As illustrated in Fig.~\ref{Fig 5: vis_levir}, our method surpasses other SOTA methods in overall detection effectiveness. In the first row of Fig.~\ref{Fig 5: vis_levir}, most models exhibit a significant number of false positives and false negatives. In contrast, our method accurately identified these change regions.

\begin{figure*}
	\centering
	\includegraphics[width=1\textwidth]{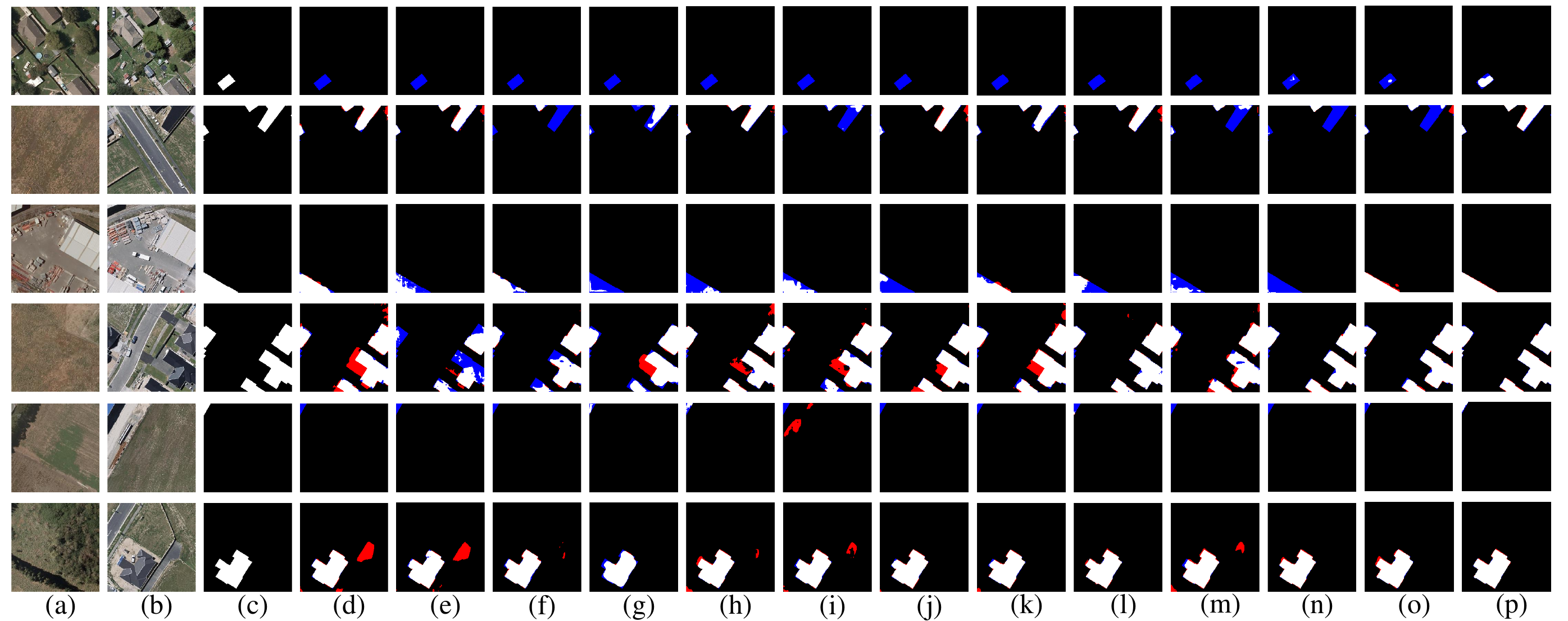}
	\caption{Visual comparative results of MMChange and other advanced methods on WHU-CD, where white indicates true positive, black indicates true negative, blue indicates false negative, and red indicates false positive. (a)T1 images, (b)T2 images, (c)Ground-truth, (d)FC-Conc, (e)FC-Diff, (f)BIT, (g)ICIF-Net, (h)ChangeFormer, (i)VcT, (j)DMINet, (k) SEIFNet, (l)BiFA, (m)SSCD-CD, (n)DGMA2Net, (o)ChangeCLIP, (p)MMChange.}
	\label{Fig 6: vis_whu}
\end{figure*}

Table ~\ref{tab1:ovaer_result} displays the experimental findings for the WHU-CD. MMChange obtained an IOU of 90.90\% and an F1 of 95.23\%, significantly outperforming the best-performing model, BiFA, among the 12 comparison models, with improvements of 1.96\% and 1.08\%, respectively. This substantial enhancement highlights the overall performance advantage of MMChange, particularly in complex change detection tasks, where it demonstrates superior adaptability and accuracy. As illustrated in Fig.~\ref{Fig 6: vis_whu}, MMChange excels in detecting large-scale building changes. Compared to other methods, it significantly reduces both missed detections and false positives in large-scale change detection, accurately capturing change features. Furthermore, MMChange also exhibits high accuracy in detecting small-scale building changes. It not only enhances detection precision but also effectively mitigates the frequent causes of missed detections and false positives encountered in other methods, showcasing its exceptional robustness and stability.

\begin{figure*}
	\centering
	\includegraphics[width=1\textwidth]{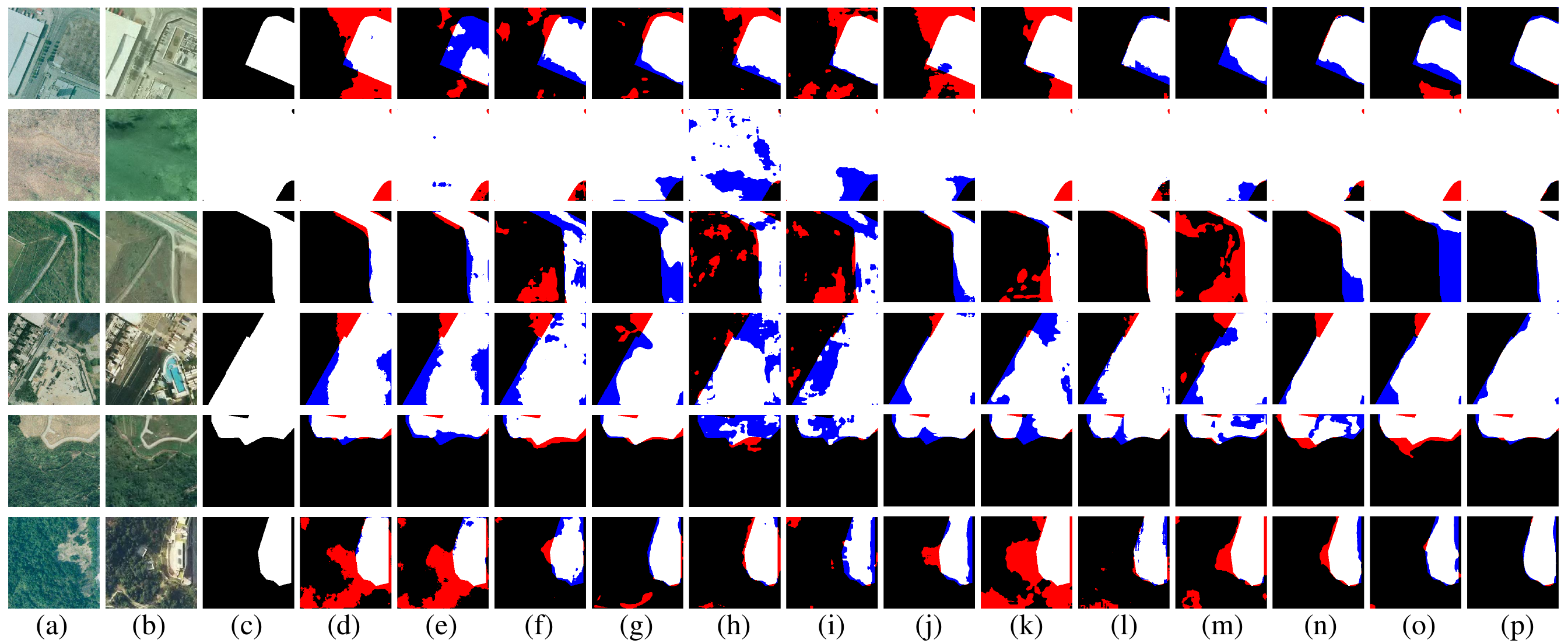}
	\caption{Visual comparative results of MMChange and other state-of-the-art methods on SYSU-CD, where white denotes true positives, black denotes true negatives, blue denotes false negatives, and red denotes false positives. (a)T1 images, (b)T2 images, (c)Ground-truth, (d)FC-Conc, (e)FC-Diff, (f)BIT, (g)ICIF-Net, (h)ChangeFormer, (i)VcT, (j)DMINet, (k)SEIFNet, (l)BiFA, (m)SSCD-CD, (n)DGMA2Net, (o)ChangeCLIP, (p)MMChange.}
	\label{Fig 7: vis_sysu}
\end{figure*}

Table ~\ref{tab1:ovaer_result} presents the experimental results for the SYSU-CD. Due to the challenging nature of the changing objects in this dataset, along with interference from noise and background elements, the detection results are susceptible to errors. Consequently, the change detection task on this dataset is particularly challenging. However, MMChange outperforms the 12 comparison models, especially in two key metrics, achieving an IOU of 72.05\% and an F1 of 83.76\%. This outcome demonstrates that, despite the increased complexity of the SYSU-CD dataset, MMChange can maintain stable and accurate performance in CD tasks. The visual results of MMChange, as illustrated in Fig.~\ref{Fig 7: vis_sysu}, indicate that it performs exceptionally well in detecting changes in buildings and land. Relative to other models, MMChange exhibits significantly fewer false positives and missed detections, effectively reducing erroneous detections in complex areas.

In summary, the experimental comparison results demonstrate that the MMChange model achieves a certain balance among parameter size, computational cost, and inference speed. Meanwhile, it attains the best performance in key accuracy metrics such as IOU and F1 across multiple datasets, including LEVIR-CD, WHU-CD, and SYSU-CD.


\subsection{Ablation Study}
\begin{figure*}
	\centering
	\includegraphics[width=1\textwidth]{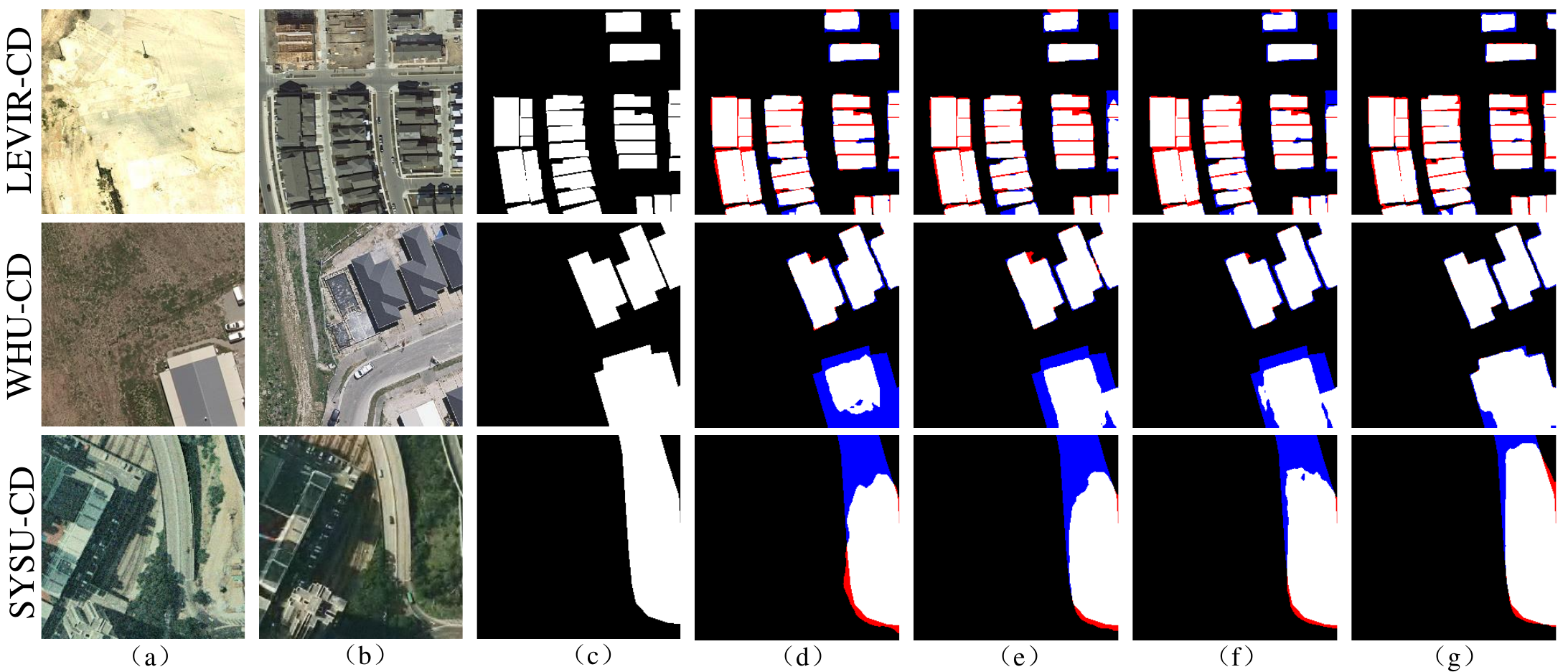}
	\caption{Visualisation of the results of different modular ablation experiments in MMChange, where white indicates true-positive, black indicates true-negative, blue indicates false-negative, red indicates false-positive, (a)T1 image, (b)T2 image, (c)Ground-truth, (d)Without IFR, (e)Without TDE, (f)Without ITFF, and (g)MMChange.}
	\label{Fig 8: vis_ablation}
\end{figure*}

In this section, An ablation study is carried out to evaluate the contribution of the text modality and the three core modules in MMChange. The results are presented in Table ~\ref{tab2: ablation_result}, and the visual outcomes are illustrated in Fig.~\ref{Fig 8: vis_ablation}. We employs 1×1 convolution and element-wise addition for dimension alignment and feature fusion.

\begin{figure*}
	\centering
	\includegraphics[width=1\textwidth]{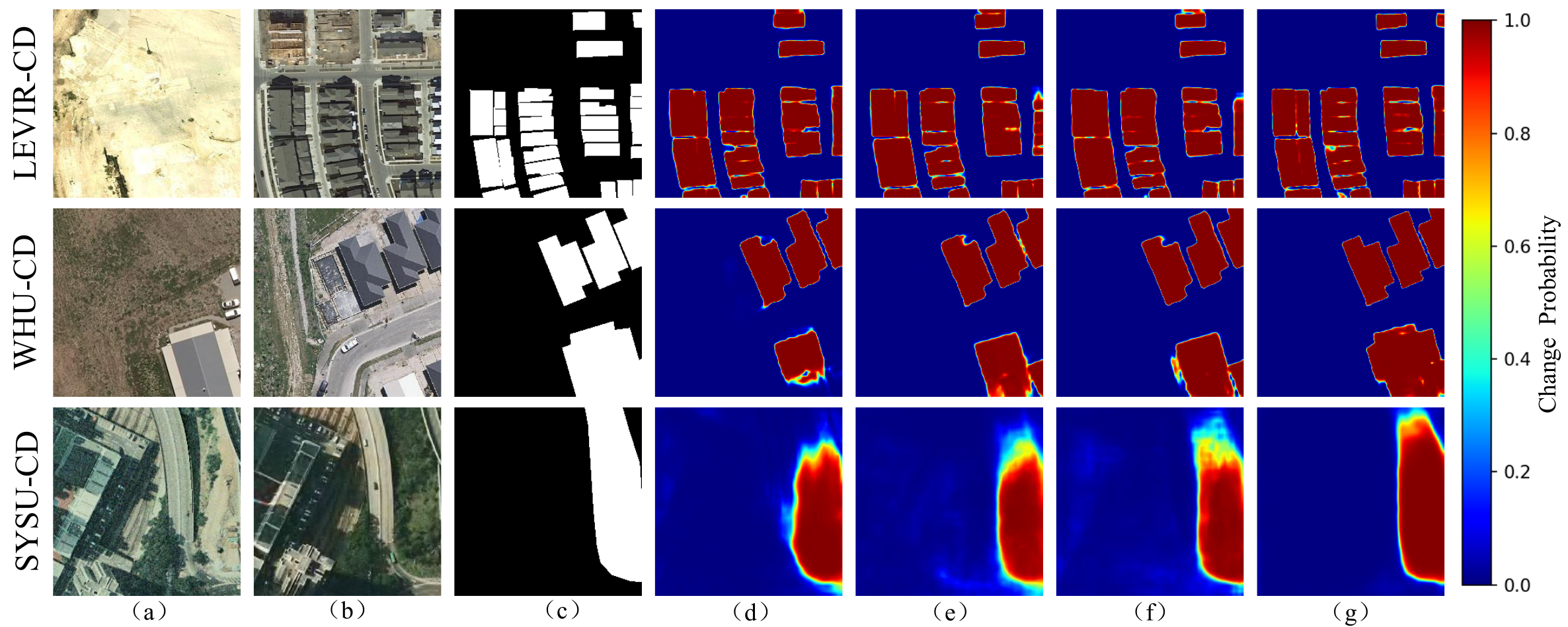}
	\caption{Heatmap results of the ablation experiments on different modules in MMChange. Red indicates areas with higher attention, while blue represents areas with lower attention. (a)T1 image, (b)T2 image, (c)Ground-truth, (d)Without IFR, (e)Without TDE, (f)Without ITFF, and (g)MMChange.}
	\label{Fig 9: vis_heat}
\end{figure*}

Text Modality Ablation: In RSCD tasks, integrating image and text modalities significantly enhances performance compared to relying solely on image modalities. This multimodal approach leverages the complementary strengths of image-based visual features and text-based semantic information, enabling a richer and more detailed contextual understanding. Table ~\ref{tab2: ablation_result} illustrates that, under the same conditions, the performance of using only image modalities is consistently lower than that of combining image and text modalities across three datasets. Specifically, the IOU scores for the unimodal approach are 80.90\%, 71.80\%, and 61.37\%, which are 2.08\%, 9.79\%, and 0.69\% lower, respectively, than those achieved by the multimodal approach. Similarly, the F1 for the unimodal approach are 89.44\%, 83.58\%, and 76.06\%, falling short by 1.26\%, 6.28\%, and 0.53\%, respectively. These results, as displayed in Table ~\ref{tab2: ablation_result}, demonstrate that incorporating text modalities into RSCD tasks substantially improves model performance.

IFR Ablation: The IFR module refines image feature information by extracting and enhancing features such as shapes and textures, thereby providing richer and more detailed information. Table ~\ref{tab2: ablation_result} displays the experimental findings of the model on the three datasets without the IFR module. Specifically, after removing the IFR, the IOU metric for without IFR is 0.88\%, 3.25\%, and 6.26\% lower than that of MMChange on the three datasets, in that order. Additionally, the F1 decreases by 0.52\%, 1.81\%, and 4.39\% on these datasets. The visual results in Fig.~\ref{Fig 8: vis_ablation} display that without the IFR module, the model's ability to detect buildings is significantly weakened, leading to an increase in false positives. For example, in the LEVIR-CD, as displayed in Fig.~\ref{Fig 8: vis_ablation}(e), compared to the model with the IFR module Fig.~\ref{Fig 8: vis_ablation}(h), more false positives (red areas) are observed. Moreover, on the SYSU-CD, as displayed in Fig.~\ref{Fig 8: vis_ablation}(e), the model without the IFR module detects fewer change areas. Both the changes in IOU and F1  from Table ~\ref{tab2: ablation_result} and the visual results in Fig.~\ref{Fig 8: vis_ablation} confirm the significant advantage of the IFR module in feature refinement.

\begin{table*}		
	\caption{Quantitative Evaluation Results Ablation Experiments For Image Encoder, Text Encoder, IFR, TDE, and ITFF On The LEVIR-CD, WHU-CD And SYSU-CD Dataset Are Presented. The Best Values Are Highlighted In Bold, And All Values Are Expressed As Percentages.
		\label{tab2: ablation_result}}
	\centering
	\renewcommand{\arraystretch}{1.5}	\resizebox{\textwidth}{!}{
		\begin{tabular}{ccccc|cccc|cccc|cccc}
			\hline
			\multirow{2}{*}{Image Encoder}& \multirow{2}{*}{Text Encoder}&\multirow{2}{*}{IFR}& \multirow{2}{*}{TDE}&\multirow{2}{*}{ITFF}                                                & \multicolumn{4}{c|}{LEVIR-CD}                                      & \multicolumn{4}{c|}{WHU-CD}                                        & \multicolumn{4}{c}{SYSU-CD}                                       \\
			\cline{6-17}
			&  &                    & &  & IOU            & F1             & R              & P              & IOU            & F1             & R              & P              & IOU            & F1             & R              & P              \\
			\hline
			\multicolumn{1}{c}{\Checkmark} &\multicolumn{1}{c}{\XSolidBrush} & \multicolumn{1}{c}{\XSolidBrush} & \multicolumn{1}{c}{\XSolidBrush}                       & \multicolumn{1}{c|}{\XSolidBrush}                        & 80.90          & 89.44          & 87.21          & 91.79          & 71.80          & 83.58          & 76.60          & 91.96          & 61.37          & 76.06          & 76.94 & 75.21          \\
			\multicolumn{1}{c}{\Checkmark} &\multicolumn{1}{c}{\Checkmark} & \multicolumn{1}{c}{\XSolidBrush} & \multicolumn{1}{c}{\XSolidBrush}                       & \multicolumn{1}{c|}{\XSolidBrush}                        & 82.98          & 90.70          & 89.90          & 92.26          & 81.59          & 89.86          & 89.10          & 90.64          & 62.06          & 76.59          & 74.31 & 79.01          \\
			\multicolumn{1}{c}{\Checkmark} &\multicolumn{1}{c}{\Checkmark} & \multicolumn{1}{c}{\XSolidBrush} & \multicolumn{1}{c}{\Checkmark}                       & \multicolumn{1}{c|}{\Checkmark}                        & 84.18          & 91.41          & 90.63          & 92.21          & 87.65          & 93.42          & 90.77          & 96.22          & 65.79          & 79.37          & \textbf{81.58} & 77.27          \\
			\multicolumn{1}{c}{\Checkmark} &\multicolumn{1}{c}{\Checkmark} &\multicolumn{1}{c}{\Checkmark} & \multicolumn{1}{c}{\XSolidBrush}                       & \multicolumn{1}{c|}{\Checkmark}                        & 84.50          & 91.60          & 90.73          & 92.49          & 88.45          & 93.87          & 90.86          & \textbf{97.08} & 68.33          & 81.19          & 77.10          & 85.73          \\
			\multicolumn{1}{c}{\Checkmark} &\multicolumn{1}{c}{\Checkmark} &\multicolumn{1}{c}{\Checkmark} & \multicolumn{1}{c}{\Checkmark}                       & \multicolumn{1}{c|}{\XSolidBrush}                        & 84.26          & 91.46          & 89.32          & \textbf{93.71} & 86.70          & 92.88          & 90.49          & 95.40          & 67.22          & 80.40          & 76.72          & 84.44          \\
			\hline
			\multicolumn{1}{c}{\Checkmark} &\multicolumn{1}{c}{\Checkmark} &\multicolumn{1}{c}{\Checkmark}                     & \multicolumn{1}{c}{\Checkmark}   & \multicolumn{1}{c|}{\Checkmark}    & \textbf{85.06} & \textbf{91.93} & \textbf{91.06} & 92.78          & \textbf{90.90} & \textbf{95.23} & \textbf{93.68} & 96.83          & \textbf{72.05} & \textbf{83.76} & 81.04          & \textbf{86.66}\\
			\hline
	\end{tabular}}
\end{table*}

TDE Ablation: The TDE module enhances the differences between bitemporal remote sensing image text descriptions generated by the LLMs, significantly boosting the model's ability to capture subtle differences. This enhancement not only focuses on local details within the text but also explores the global relationships between the texts. Furthermore, by incorporating the text modality, the TDE module effectively compensates for the limitations of relying solely on image-only features, thereby boosting the overall performance and robustness of the model. Table ~\ref{tab2: ablation_result} presents the experimental findings of the model on the three datasets without the TDE. Specifically, after removing the TDE module, the IOU metric for without TDE is 0.56\%, 2.45\%, and 3.72\% lower than that of MMChange on the three datasets, in that order. Additionally, the F1 decreases by 0.33\%, 1.36\%, and 2.57\% on these datasets. The visual results in Fig.~\ref{Fig 8: vis_ablation} display that without the TDE module, the model's ability to detect building and land changes is significantly weakened, leading to an increase in missed detections. For example, on the WHU-CD, as demonstrated in Fig.~\ref{Fig 8: vis_ablation}(f), compared to the model with the TDE module Fig.~\ref{Fig 8: vis_ablation}(h), more missed change information (blue areas) can be observed. Moreover, on the LEVIR-CD, as displayed in Fig.~\ref{Fig 8: vis_ablation}(f), the model without the TDE module has a greater number of falsely detected change regions (red areas). Both the quantitative analysis of IOU and F1  in Table ~\ref{tab2: ablation_result} and the visual displays in Fig.~\ref{Fig 8: vis_ablation} prove the effectiveness of the TDE module in boosting the model's performance.

\begin{table*}
	\caption{Quantitative Evaluation Results of Ablation Experiments with Different Backbones on the LEVIR-CD, WHU-CD, and SYSU-CD Datasets with the Best Values Highlighted in Bold, and All Values Expressed in Percentage
		\label{tab3: Backbone_result}}
	\centering
	\tiny
	\renewcommand{\arraystretch}{1.2}\resizebox{\textwidth}{!}{
		\begin{tabular}{c|cccc|cccc|cccc}
			\hline
			\multirow{2}{*}{Method} & \multicolumn{4}{c|}{\textbf{LEVIR-CD}}                                                           & \multicolumn{4}{c|}{\textbf{WHU-CD}}                                                     
			& \multicolumn{4}{c}{\textbf{SYSU-CD}} \\
			\cline{2-13}
			& \multicolumn{1}{c}{IOU} & \multicolumn{1}{c}{F1} & \multicolumn{1}{c}{R} & \multicolumn{1}{c|}{P} & \multicolumn{1}{c}{IOU} & \multicolumn{1}{c}{F1} & \multicolumn{1}{c}{R} & \multicolumn{1}{c|}{P} & \multicolumn{1}{c}{IOU} & \multicolumn{1}{c}{F1} & \multicolumn{1}{c}{R} & \multicolumn{1}{c}{P}\\
			\hline          
			ResNet18 & 84.44          & 91.57          & 90.78          & 92.36          & 90.30          & 94.90          & 93.36          & 96.50          & 71.88          & 83.64          & 81.42 & 85.99          \\
			ResNet34 & 84.35          & 91.51          & \textbf{91.16} & 91.87          & 89.01          & 94.18          & 93.61          & 94.76          & 71.82          & 83.60          & \textbf{83.23}          & 83.98          \\
			\hline
			ResNet50 & \textbf{85.06} & \textbf{91.93} & 91.06          & \textbf{92.78} & \textbf{90.90} & \textbf{95.23} & \textbf{93.68} & \textbf{96.83} & \textbf{72.05} & \textbf{83.76} & 81.04          & \textbf{86.66}\\
			\hline
	\end{tabular}}
\end{table*}

\begin{table*}
	\caption{Quantitative Evaluation Results Of Variability Experiments On LEVIR-CD, WHU-CD AND SYSU-CD Under Noisy Or Lighting Conditions
		\label{tab8+:Variabilities }}
	\centering
	\renewcommand{\arraystretch}{1.2}\resizebox{\textwidth}{!}{
	\scriptsize
	\begin{tabular}{c|cccc|cccc|cccc}
		\hline
		\multirow{2}{*}{Method} &\multicolumn{4}{c|}{\textbf{LEVIR-CD}}                                                           & \multicolumn{4}{c|}{\textbf{WHU-CD}}                                                     
		& \multicolumn{4}{c}{\textbf{SYSU-CD}}                                                             \\
		\cline{2-13}
		& \multicolumn{1}{c}{IOU} & \multicolumn{1}{c}{F1} & \multicolumn{1}{c}{R} & \multicolumn{1}{c|}{P} & \multicolumn{1}{c}{IOU} & \multicolumn{1}{c}{F1} & \multicolumn{1}{c}{R} & \multicolumn{1}{c|}{P} & \multicolumn{1}{c}{IOU} & \multicolumn{1}{c}{F1} & \multicolumn{1}{c}{R} & \multicolumn{1}{c}{P}  \\
		\hline
		baseline             & 80.90          & 89.44          & 87.21          & 91.79          & 71.80          & 83.58          & 76.60          & 91.96          & 61.37          & 76.06          & 76.94 & 75.21          \\
		baseline+noise       & 79.47                    & 88.56                  & 86.81                 & 90.38                 & 70.60                   & 82.76                  & 73.79                 & 94.22                &  61.07                  & 75.83                   &  78.47                & 73.37                  \\
		baseline+light       & 79.87                    & 88.81                  & 86.89                 & 90.81                 &69.75                     &82.18                   & 79.97                  & 84.51                 & 56.11                   &  71.89                 &  76.15                &  68.07                    \\
		baseline+noise+light & 78.52                   & 87.97                  & 86.29                 & 89.71                 & 65.78                   & 79.35                  & 84.65                 & 74.68                  & 53.76                   & 69.93                  & 71.97                 & 68.00                     \\
		Ourst+noise       & 83.00                   & 90.71                  & 89.23                 & 92.25                 & 78.26                   & 87.80                  & 83.68                 & 92.35                 & 71.12                   & 83.12                   & \textbf{83.74}                 & 82.52                \\
		Ours+light       & 83.31                   & 90.90                  & 89.53                 & 92.30                 & 81.98                   & 90.10                  & 86.51                 & 94.00                 & 71.04                   & 83.07                   & 82.91                 & 83.22                \\
		Ours+noise+light & 82.45                   & 90.38                  & 89.08                 & 91.72                 & 77.03                   & 87.02                  & 79.66                  & 95.89                 & 70.32                   & 82.58                  & 82.74                 & 82.42                            \\
		\hline
			Ours                    & \textbf{85.06} & \textbf{91.93} & \textbf{91.06} & \textbf{92.78}          & \textbf{90.90} & \textbf{95.23} & \textbf{93.68} & \textbf{96.83} & \textbf{72.05} & \textbf{83.76} & 81.04          & \textbf{86.66} \\
		\hline                  
	\end{tabular}}
\end{table*}

\begin{table*}
	\caption{Quantitative Evaluation Results Using Different Prompts and VLM on LEVIR-CD, WHU-CD, and SYSU-CD
		\label{tab9+:Variabilities }}
	\centering
	\renewcommand{\arraystretch}{1.2}\resizebox{\textwidth}{!}{
		\scriptsize
		\begin{tabular}{c|cccc|cccc|cccc}
			\hline
			\multirow{2}{*}{Method} &\multicolumn{4}{c|}{\textbf{LEVIR-CD}}                                                           & \multicolumn{4}{c|}{\textbf{WHU-CD}}                                                     
			& \multicolumn{4}{c}{\textbf{SYSU-CD}}                                                             \\
			\cline{2-13}
			& \multicolumn{1}{c}{IOU} & \multicolumn{1}{c}{F1} & \multicolumn{1}{c}{R} & \multicolumn{1}{c|}{P} & \multicolumn{1}{c}{IOU} & \multicolumn{1}{c}{F1} & \multicolumn{1}{c}{R} & \multicolumn{1}{c|}{P} & \multicolumn{1}{c}{IOU} & \multicolumn{1}{c}{F1} & \multicolumn{1}{c}{R} & \multicolumn{1}{c}{P}  \\
			\hline
			Prompt1             & 84.85          & 91.80          & \textbf{91.12}          & 92.49          & 81.80          & 89.99          & 85.61          & 94.84          & 71.07          & 83.09          & \textbf{84.74} & 81.49          \\
			Prompt2       & 84.79                    & 91.77                  & 90.40                 & 93.18                 & 82.87                   & 90.63                  & 90.88                 & 90.39                &  71.19                  & 83.17                   &  80.61                & 85.90                  \\
			Prompt3       & 84.83                    & 91.80                  & 90.53                 & 93.10                 &85.02                     &91.91                   & 86.54                  & \textbf{97.99}                 & 71.54                   &  83.41                 &  83.57                &  83.24                    \\
			Prompt4 & 84.69                   & 91.71                  & 90.26                 & \textbf{93.21}                 & 83.16                   & 90.81                  & 84.90                 & 97.60                  & 70.97                   & 83.02                  & 79.53                 & 86.83                     \\
			CLIP Interrogator       & 84.50                   & 91.60                  & 91.00                 & 92.20                 & 88.17                   & 93.71                  & 92.09                 & 95.39                 & 71.46                   & 83.35                   & 81.87                 & 84.90                \\
			mPLUG       & 84.65                   & 91.69                  & 90.88                 & 92.52                & 87.61                   & 93.40                  & 91.71                 & 95.15                & 71.11                   & 83.12                   & 79.42                 & \textbf{87.17}                \\
			\hline
			Ours                    & \textbf{85.06} & \textbf{91.93} & 91.06 & 92.78          & \textbf{90.90} & \textbf{95.23} & \textbf{93.68} & 96.83 & \textbf{72.05} & \textbf{83.76} & 81.04          & 86.66 \\
			\hline                  
	\end{tabular}}
\end{table*}
ITFF Ablation: The ITFF enhances the model's ability to capture both image and text feature by combining different types of attention mechanisms. This design not only facilitates the comprehensive extraction and understanding of key information from multimodal data but also effectively integrates this information for efficient multimodal feature fusion. Table ~\ref{tab2: ablation_result} displays the experimental results on the three datasets without the ITFF. Specifically, compared to MMChange, the IOU metric for without ITFF is 0.80\%, 4.20\%, and 4.83\% lower, and the F1 decreases by 0.47\%, 2.35\%, and 3.36\% on these datasets, respectively. The visual results in Fig.~\ref{Fig 8: vis_ablation} further display that the absence of the ITFF module leads to poor performance in detecting building and land changes, with an increase in missed detections. For example, on the SYSU-CD, as displayed in Fig.~\ref{Fig 8: vis_ablation}(g), compared to the model with the ITFF Fig.~\ref{Fig 8: vis_ablation}(h), more missed change areas (blue regions) are clearly observed. Moreover, on the LEVIR-CD, as displayed in Fig.~\ref{Fig 8: vis_ablation}(g), the model without the ITFF module also exhibits missed detections when handling dense building changes. The quantitative analysis of IOU and F1 in Table ~\ref{tab2: ablation_result}, along with the visual displays in Fig.~\ref{Fig 8: vis_ablation}, strongly supports the role and effectiveness of the ITFF module in improving the model's performance.

Additionally, we carried out a heatmap comparison of the ablation experiments on various modules in MMChange, as illustrated in Fig.~\ref{Fig 9: vis_heat}. The heatmaps in Fig.~\ref{Fig 9: vis_heat} clearly demonstrate that as the IFR, TDE, and ITFF modules are progressively integrated, the model's attention becomes increasingly concentrated on the areas of change. In the absence of these modules, the change areas in the heatmap appear more dispersed, resulting in a less distinct focus on the changes and leads to more false positives and missed detections. However, after incorporating the IFR, TDE, and ITFF modules, the heatmap reveals a significant improvement in the model's response to the change areas, enabling it to more accurately concentrate on the regions of change.
\subsection{Discussion}

Discussion on Different Backbones: To strengthen the validation of MMChange's performance, we carried out experiments with various backbone networks, including ResNet50, ResNet34, and ResNet18. The results displays in Table ~\ref{tab3: Backbone_result} indicate that our method outperforms the models utilizing ResNet18 and ResNet34 in terms of both IOU and F1 across the three datasets. This further underscores the superior performance of MMChange in accurately detecting areas of change, emphasizing its significant advantage in RSCD tasks.
\begin{figure*}
	\centering
	\includegraphics[width=1\textwidth]{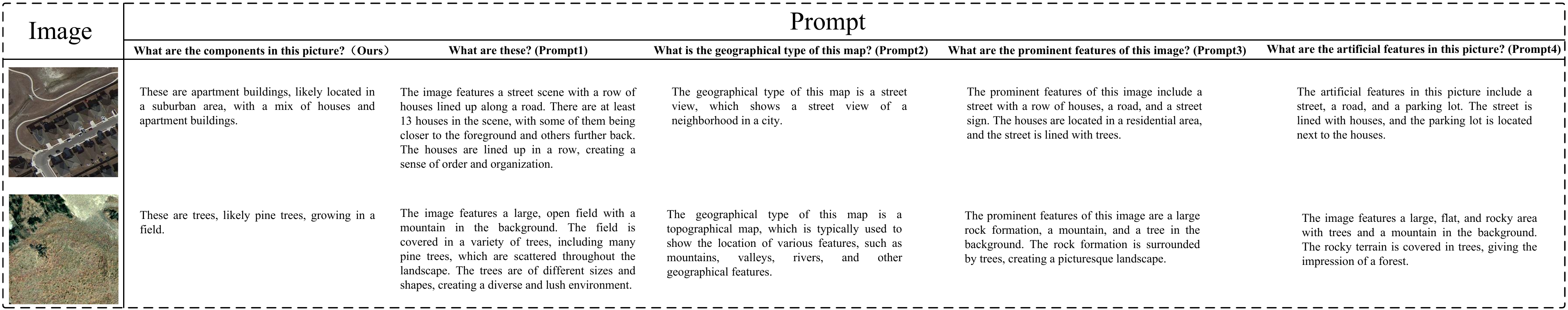}
	\caption{Text Descriptions of Bitemporal Remote Sensing Images Generated by TinyLLaVA with Different Prompts}
	\label{Fig 10: vis_prompt}
\end{figure*}
Variabilities Experiment: Inspired by \cite{hong2018augmented}, we manually introduced noise and illumination variations into three datasets to further evaluate the robustness and performance of MMChange under different variability conditions. The experimental results are presented in Table~\ref{tab8+:Variabilities }. As indicated in the table, MMChange outperforms the baseline model in terms of IOU by 3.93\%, 11.25\%, and 16.56\% across the three datasets, respectively. Additionally, it achieves improvements in the F1 of 2.41\%, 7.67\%, and 12.65\%. These results demonstrate that MMChange significantly enhances change detection performance in the presence of noise and illumination variations, showcasing strong resistance to interference.

Analysis of Prompt Design and VLM: To evaluate the effectiveness of prompt engineering and the appropriateness of adopting TinyLLaVA as the text generation model, we designed four distinct prompts and employed them to produce bitemporal image descriptions using TinyLLaVA, as shown in Fig.~\ref{Fig 10: vis_prompt}. As reported in Table~\ref{tab9+:Variabilities }, although these prompts do not surpass the performance of the MMChange-specific prompt “What are the components in this picture?”, they still achieve superior results compared to most unimodal baselines listed in Table~\ref{tab1:ovaer_result}. This indicates that incorporating textual features, when guided by well-crafted prompts, contributes positively to model performance by providing richer semantic cues beyond those captured by visual features alone. Furthermore, we compared TinyLLaVA with two alternative VLM (CLIP Interrogator\cite{jagtap2025image} and mPLUG\cite{li2022mplug}) under the same prompt setting. As presented in Table~\ref{tab9+:Variabilities }, experimental results on the WHU-CD dataset show that TinyLLaVA yields higher F1 and IOU, with improvements of 1.53\% and 2.73\% over CLIP Interrogator, and 1.83\% and 3.29\% over mPLUG, respectively. These findings highlight the advantages of TinyLLaVA in generating semantically aligned and task-relevant descriptions, thereby enhancing the overall effectiveness of multimodal remote sensing change detection.

\section{Conclusion}
This paper introduces MMChange, a multi-modal remote sensing change detection (RSCD) model built upon VLM. Unlike most existing single-modal RSCD models, MMChange not only utilizes image modality information but also integrates text modality information, which presents both challenges and significant importance. To solve this, we designed an IFR module to enhance the detection of change areas by refining image features. Subsequently, we proposed a TDE module that emphasizes the key features of the change areas. It achieves this by highlighting the differences between the text descriptions of bitemporal RS images, enabling the model to better focus on these change regions. Furthermore, to merge multi-modal information, we developed an ITFF module that efficiently combines image and text features, fully leveraging the complementary strengths of both modalities, which in turn improves detection accuracy and robustness. We performed experiments on three commonly used datasets to assess the effectiveness of MMChange. The results prove that MMChange outperforms several advanced models, achieving superior detection outcomes. Additionally, we performed ablation experiments to thoroughly assess the effectiveness of each module within MMChange. Looking ahead, although MMChange has achieved significant progress in multimodal RSCD, its reliance on high-quality annotated data limits its applicability in data-scarce scenarios. In the next phase of research, we plan to explore the use of VLM to automatically generate labels for remote sensing images, reducing dependence on manual annotation. Leveraging the text generation capabilities of VLM could advance the development of multimodal self-supervised, weakly supervised, and unsupervised methods, consequently enhancing both the accuracy and efficiency of RSCD.

\bibliographystyle{IEEEtran}
\bibliography{second_tgrs}

\vfill

\end{document}